\documentclass[11pt,a4paper]{article}
\usepackage{times}
\usepackage{latexsym}
\usepackage{comment}
\usepackage[]{acl}

\usepackage{microtype}

\usepackage[linesnumbered, ruled]{algorithm2e}
\usepackage[T1]{fontenc}
\usepackage{dsfont}

\usepackage{mdframed}
\usepackage{booktabs}
\usepackage{graphicx}
\usepackage{commath}

\usepackage[normalem]{ulem}
\usepackage{enumitem}
\usepackage{xcolor}
\usepackage{soul}
\colorlet{soulred}{red!30}
\sethlcolor{soulred}%

\usepackage{amsmath,amsthm,amsfonts,amssymb,bm}
\usepackage{framed}
\usepackage{varioref}
\usepackage{float}
\usepackage{lipsum}
\usepackage{multirow}
\usepackage{dashrule}
\usepackage{url}
\usepackage{pifont}

\definecolor{torange}{HTML}{FE6100}
\definecolor{tpurple}{HTML}{785EF0}
\definecolor{tpink}{HTML}{DC267F}

\usepackage{array}

\title{Automatic Error Analysis \\ for Document-level Information Extraction}

\author{Aliva Das$^*$, Xinya Du$^*$, Barry Wang$^*$,\\
\textbf{Kejian Shi, Jiayuan Gu, Thomas Porter, Claire Cardie}\\
Department of Computer Science, Cornell University \\
{\tt \{ad677, xd75, zw545, ks2325, jg844, tjp78, ctc9\}@cornell.edu}
}

\date{}

\begin{document}
\maketitle
\def\thefootnote{*}\footnotetext{These authors contributed equally to this work.}\def\thefootnote{\arabic{footnote}}

\begin{abstract}
Document-level information extraction (IE) tasks have recently begun
to be revisited in earnest using the end-to-end neural network techniques
that have been successful on their sentence-level IE counterparts.
Evaluation of the approaches, however, has been limited  in a number of dimensions.
In particular, the precision/recall/F1 scores typically reported 
provide few insights on the range of errors the models make. 
We build on the work of \newcite{kummerfeld-klein-2013-error} 
to propose a transformation-based framework 
for automating error analysis in document-level event and (N-ary) 
relation extraction.
We employ our framework to compare two state-of-the-art document-level 
template-filling approaches on datasets from three domains;
and then, to gauge progress in IE since its inception 30 years ago, 
vs.\ four systems from the~\newcite{muc-1992-message} evaluation.\footnote{Our code for the error analysis tool and its output on different model predictions are available at \url{https://github.com/IceJinx33/auto-err-template-fill/.}}
\end{abstract}

\section{Introduction}
\label{sec:intro}
Although information extraction (IE) research has
almost uniformly focused on \textit{sentence-level} relation and event extraction \cite{grishman2019twenty},
the earliest research in the area formulated the task at the \textit{document level}.  Consider, for example,
the first large-scale (for the time) evaluations of IE 
systems \phantom{  }--- e.g.\ \newcite{muc-1991-message} and~\newcite{muc-1992-message}.  Each involved a complex document-level event
extraction task: there were 24 types of events, over a dozen event arguments (or \textit{roles}) to be
identified for each event;
documents could contain zero to tens of events, and extracting argument
entities (or \textit{role fillers}) required noun phrase coreference 
resolution to ensure interpretability for the end-user (e.g.\ to ensure that
multiple distinct mentions of the same entity
in the output were not misinterpreted as references to distinct entities).

\begin{figure*}[t]
\centering
\resizebox{\textwidth}{!}{
\includegraphics{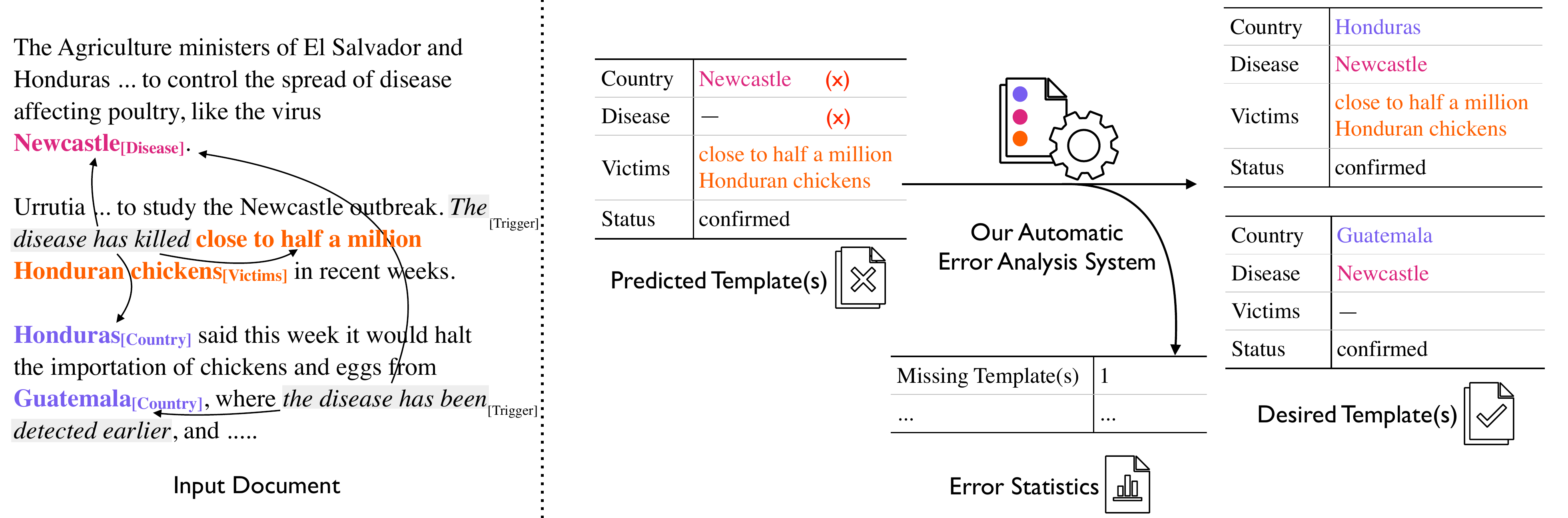}
}
\caption{The document-level extraction task from the ProMED dataset on disease outbreaks (left) and the automatic error analysis process (right). Our system performs a set of transformations on the predicted templates to convert them into the corresponding gold standard templates. Transformation steps are mapped to corresponding error types to produce informative error statistics.} 
\label{fig:task}
\end{figure*}

The task was challenging: information relevant for a single event
could be scattered across the document or repeated in multiple places; relevant information might need to be shared across multiple events; information regarding different events could be intermingled. 
In Figure \ref{fig:task}, for example, the  {\sc Disease} "Newcastle" is mentioned
well before its associated event is mentioned (via the triggering phrase "the disease has killed"); that same mention of "Newcastle" must again be recognized as the {\sc Disease} in a second event; and the \textsc{Country} of the first event ("Honduras") appears only in the sentence describing the second event.

In fact, the problem of document-level information extraction has only
recently begun to be revisited 

\cite{quirk-poon-2017-distant,jain-etal-2020-scirex, du-etal-2021-template, du2021grit,
li2021documentlevel, du2021towards, yang-etal-2021-document} in part in an attempt to test the power of end-to-end neural
network techniques that have been so successful on their sentence-level counterparts.\footnote{See, for example, \newcite{Zhang2019ExtractingEA}, \newcite{du-cardie-2020-event} and
\newcite{lin-etal-2020-joint} for within-sentence event extraction; \newcite{akbik-etal-2018-contextual}, and \newcite{akbik-etal-2019-pooled} for named entity recognition (NER); and \newcite{zhang-etal-2018-graph} and \newcite{luan-etal-2019-general} for sentence-level relation extraction.}
Evaluation, however, 
has been limited in a number of ways. 
First, despite the relative complexity of the task, approaches are only evaluated 
with respect to their overall performance
scores (e.g.\ precision, recall, and F1). Even though scores at the role level are sometimes included, no systematic analysis or characterization of the types 
of errors that occur is typically done.  The latter is needed to 

determine strategies to improve performance, to obtain more informative
cross-system and cross-genre comparisons, and to identify and track broader
advances in the field as the underlying approaches evolve.
To date, for example, 
there has been no attempt to directly compare the error landscape and distribution of
newly developed neural IE methods with that of the largely hand-crafted systems of the 1990s.

\begin{figure*}[hbt!]
\centering
\resizebox{\textwidth}{!}{
\includegraphics{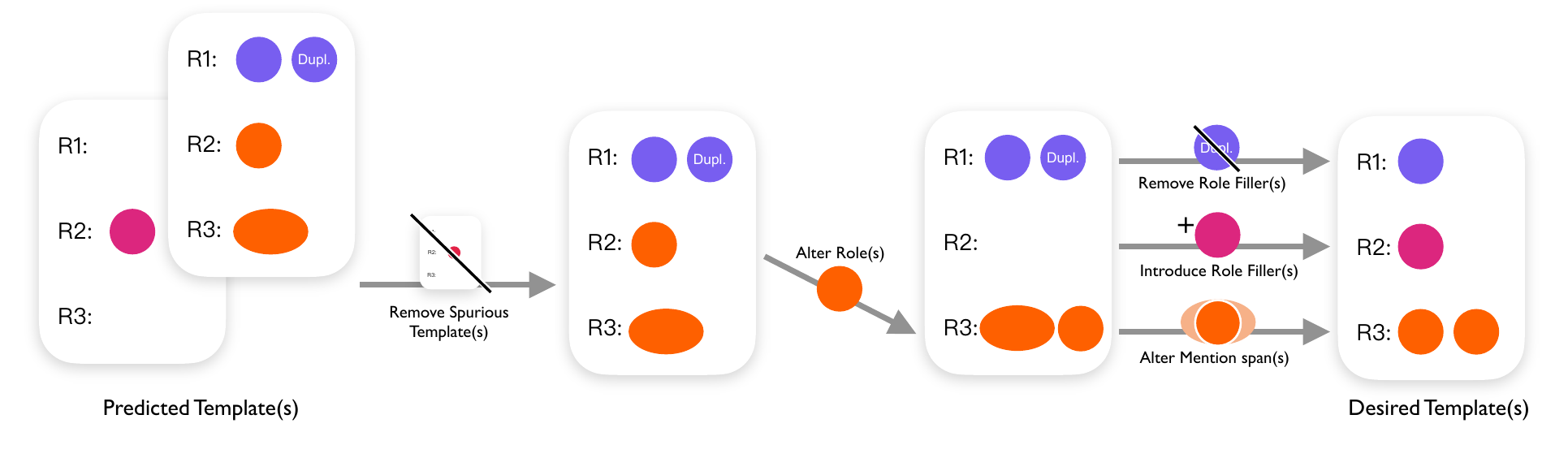}
}
\caption{Automatic transformations to convert predicted templates (on the left) to gold templates (on the right). Arrows represent transformations. Colored circles represent role filler entity mentions. \textit{Dupl.} stands for duplicate.}
\label{fig:transformation}
\end{figure*}

In this work, we first introduce a framework for automating error
analysis for document-level event and relation extraction, casting both as
instances of a general role-filling, or \textit{template-filling task} 
\cite{JurafskyandMartin3rdChap17}. Our approach converts predicted system outputs
into their gold standard counterparts through a series
of template-level transformations (Figure~\ref{fig:transformation}) and then maps
combinations of transformations into a collection of IE-based error types.
Examples of errors include duplicates, missing and spurious role fillers, missing and spurious templates, and incorrect role and template assignments for fillers. (See Figure~\ref{fig:error_types} for the full set).

Next, we employ the error analysis framework in a comparison of two
state-of-the-art document-level neural template-filling approaches, DyGIE++ \cite{wadden-etal-2019-entity} and GTT \cite{du-etal-2021-template},
across three template-filling datasets (SciREX, ProMED \cite{patwardhan-riloff-2009-unified}\footnote{\url{http://www.promedmail.org}}, and MUC-4).
 
Finally, in an attempt to gauge progress in the information
extraction field over the past 30 years, we employ the framework to compare
the performance of four of the original MUC-4 systems with the two newer deep-learning
approaches to document-level IE.\footnote{The 1992 model outputs are available in the
MUC-4 dataset released by NIST, available at
\url{https://www-nlpir.nist.gov/related_projects/muc/muc_data/muc_data_index.html.}}

We find that (1)  the best of the early IE models --- which strikes a better balance between precision and recall --- outperforms modern models that exhibit much higher precision and much lower recall; (2) the  modern neural models make more mistakes on scientific vs.\ news-oriented texts, and missing role fillers is universally the largest source of errors; and (3) modern models have clear advantages over the early IE systems in terms of accurate span extraction, while the early systems make fewer mistakes assigning role fillers to their roles.

\section{Related Work}
Aside from the original MUC-4 evaluation scoring reports \cite{Chinchor92}, which included counts of missing and spurious role filler errors, there have been very few attempts at understanding the types of errors made by IE systems and grounding those errors linguistically. \newcite{7489041} proposed a framework for studying how different errors propagate through an IE system; however, the framework can only be used for pipelined systems, not end-to-end ones.

On the other hand, automated error analysis with linguistically motivated error types has been used in other sub-fields of NLP such as machine-translation \cite{vilar-etal-2006-error, zhou-etal-2008-diagnostic, farrus-etal-2010-linguistic, auto-err-morpho-rich, Zeman2011AddicterWI, popovic-ney-2011-towards}, coreference resolution \cite{uryupina-2008-error, kummerfeld-klein-2013-error, martschat-strube-2014-recall, martschat2015analyzing} and parsing~\cite{kummerfeld-etal-2012-parser}. 
Recently, generalized automated error analysis frameworks involving human-in-the-loop testing like Errudite \cite{wu2019errudite},  
\textsc{CheckList} \cite{ribeiro-etal-2020-beyond}, 
CrossCheck \cite{arendt-etal-2021-crosscheck}, and AllenNLP Interpret \cite{DBLP:journals/corr/abs-1909-09251} have successfully been applied to tasks like machine comprehension and relation extraction~\cite{DBLP:journals/corr/abs-2004-14855}.
Closest to our work are \newcite{kummerfeld-etal-2012-parser} and \newcite{kummerfeld-klein-2013-error}, which use model-agnostic transformation-based mapping approaches to automatically obtain error information in the predicted structured output.

\section{Template-Filling Task Specification and Evaluation}
\label{sec:taskdef}

As in \newcite{JurafskyandMartin3rdChap17}, we will refer to
document-level information extraction tasks as \textit{template-filling tasks} and use the term
going forward to refer to both event extraction and document-level relation extraction tasks.

Given a document, $D$, and an IE template specification

consisting of a predetermined list of roles $R_1, R_2, ..., R_i$ associated with each type of
relevant event for the task of interest, the goal for template filling is to extract from $D$, one output template, $T$ for every relevant event/relation $e_1, e_2, \ldots, e_n$ present in the document. 
Notably, in the general case, $n \geq 0$ and is not specified as part of the input.
In each output template, its roles are filled with the corresponding role filler(s), which can be inferred or extracted from the document depending on the predetermined role types. We consider two role types here:\footnote{There 
are potentially more role types depending on the dataset (e.g.\ normalized dates, times, locations); we will not consider those here.}

\textbf{Set-fill roles}, which must be filled with exactly one role filler from a finite set supplied in the template specification. An example of a set-fill role in Figure~\ref{fig:task} is \textsc{Status}, which can be {\tt confirmed}, {\tt possible}, or {\tt suspected}.

\textbf{String-fill roles}, whose role filler(s) are spans extracted from the document, or left empty if no 
corresponding role filler is found in the document.  \textsc{Victims}, \textsc{Disease} and \textsc{Country} are string-fill roles in Figure~\ref{fig:task}.
Some string-fill roles allow multiple fillers; for example, there might be more than one \textsc{Victims}.
Importantly, for document-level template filling,
exactly one string should be included for each role filler entity (typically a canonical mention of the
entity), i.e.\ coreferent mentions of the same entity are not permitted.

\paragraph{Evaluation.} We use the standard (exact-match) F1 score \cite{Chinchor92} to evaluate the output produced by a template-filling system: 
\begin{equation}
\nonumber
    F1 = \frac{2\cdot \text{Precision} \cdot \text{Recall}}{\text{Precision}+\text{Recall}}
\end{equation}

\section{Methodology: Automatic Transformations for Error Analysis}

Similar to the work of \newcite{kummerfeld-klein-2013-error}, our error analysis approach is system-agnostic, i.e.\ it only uses system output and does not consider intermediate system decisions. This allows for error analysis and comparison across different kinds of systems --- end-to-end or pipeline; neural or pattern-based.

Given inputs consisting of the system-predicted templates and gold standard templates (i.e.\ desired output) for every document in the target dataset, our error analysis tool operates in three steps.  For each document,
\begin{enumerate}
    \item Perform an {\it
optimized mapping} of the associated predicted templates and gold templates.

   \item Apply a pre-defined set of {\it transformations} to convert each system-predicted
   template into the desired gold template, keeping track of the transformations applied.
   
   \item Map the changes made in the conversion process to an IE-based set of {\it error types}.
\end{enumerate}
\noindent
We describe each step in detail in the subsections below.

\subsection{Optimized Matching}

The first stage of the error analysis tool involves matching each system-predicted template to the best-matching gold template for each document in the dataset. In particular, the overall F1 score for a given document can vary based on how a predicted template is individually matched with a gold template (or left unmatched). 

Specifically, for each document, we recursively generate all possible {\it template matchings} --- where each predicted template is matched (if possible) to a gold template. 
In particular, for a document with $P$ predicted templates and $G$ gold templates, the total number of possible template matchings is: 
\begin{equation*}
\small
\resizebox{\columnwidth}{!}{
$\begin{gathered}
    1 + {P \choose 1}G + {P \choose 2}G(G-1) + ... + \frac{G!}{(G-P)!}, \text{if } G - P \geq 0 \\
    1 + {P \choose 1}G + {P \choose 2}G(G-1) + ... + {P \choose G}G!, \text{if } G - P < 0 \\
    =  \sum_{i=0}^{min(P, G)} {P \choose i} \frac{G!}{(G-i)!}
\end{gathered}$}
\end{equation*}
Note that template matching can result in unmatched predicted templates ({\it Spurious Templates}), as well as unmatched gold templates ({\it Missing Templates}).

Next, for each predicted-gold pair in a template matching, we iterate through all its roles and recursively generate all possible {\it mention matchings}, in each of which a predicted role filler is matched (if possible) to a set of coreferent gold role fillers. 
Similar to template matching, the process of mention matching can also result in unmatched predicted role fillers ({\it Spurious Role Fillers}) and unmatched coreferent sets of gold role fillers ({\it Missing Role Fillers}).

Through the process, each predicted role filler increases the denominator of the total precision by 1, and each set of coreferent gold role fillers increases the denominator of total recall by 1. Whenever there is a matched mention pair in which the predicted role filler has an exact match to an element of the set of coreferent gold role fillers, this adds 1 to the numerator of both precision and recall. These counts are calculated for each template matching.

Using precision and recall, the total F1 score across all the slots/roles is calculated and the template matching with the highest total F1 score is chosen. If there are ties, the template matching with the fewest errors is chosen (see Section~\ref{subsec:error-type-mappings}).

\begin{figure*}[!ht]
\centering
\resizebox{\textwidth}{!}{
\includegraphics{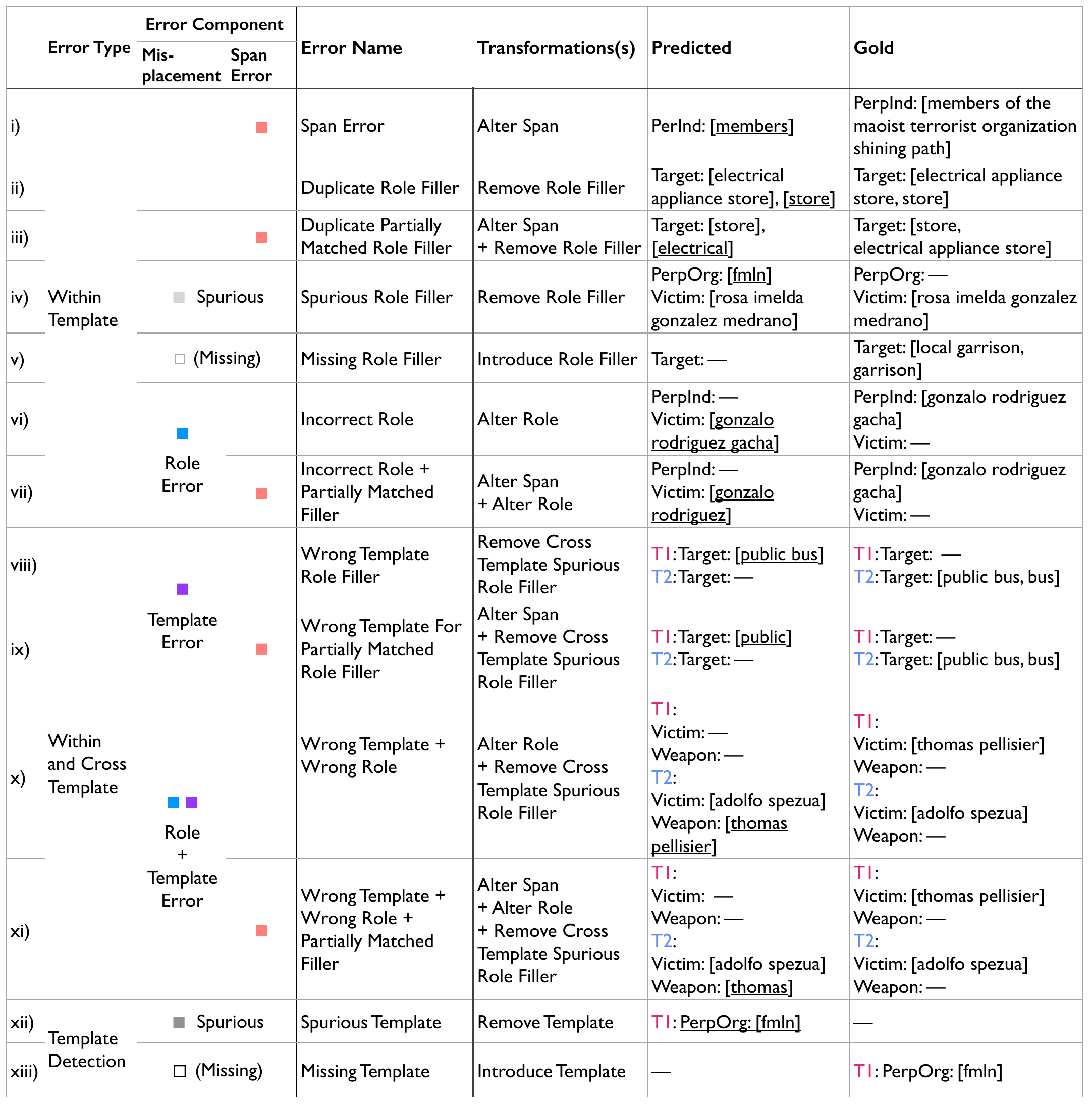}
}
\caption{Error Types with examples from the MUC-4 dataset. For each template, in every role, the role fillers in brackets refer to the same entity, while role fillers in different brackets refer to different entities. The underlined text indicates the error in the prediction.}
\label{fig:error_types}
\end{figure*}

\subsection{Transformations}
\label{sec:transformation}

The second part of the error analysis tool involves changing the predicted templates to the desired gold templates with the help of a fixed set of transformations as detailed below.

\begin{itemize}
    \item  {\bf Alter Span} transforms a role filler into the gold role filler with the lowest {\it span comparison score} ($SCS$). The tool provides two options 
    for computing the SCS between two spans, and each depends
    only on the starting and ending indices of the spans.\footnote{This deviates from \newcite{kummerfeld-klein-2013-error}, in which incorrect spans are altered to gold mentions that have the same head token, requiring the use of a syntactic parser.} 
    SCS can be interpreted as distance and is 0 between two identical spans, and 1 for non-overlapping spans. The two modes are given as follows:

a) {\it absolute}: 
This mode captures the (positive) distance between the starting (and ending) character offsets of spans $x$ and $y$ in the document, and scales that value by the sum of the lengths of $x$ and $y$, capping it at a maximum of 1.

\begin{equation}
\nonumber
\resizebox{0.9\columnwidth}{!}{
$SCS = max\left(1, \frac{\abs{x_{start} - y_{start}} + \abs{x_{end} - y_{end}}}
{\text{length}(x) + \text{length}(y)}\right)$
}
\end{equation}
b) {\it geometric mean}:

This mode captures the degree of disjointedness between spans $x$ and $y$ by dividing the length of the overlap between the two spans with respect to each of their lengths, multiplying those two fractions, and subtracting the final result from 1. 

If $si$ is the length of the intersection of $x$ and $y$, and 
neither $x$ nor $y$ have length 0, $SCS$ is calculated as shown below; otherwise, $SCS$ is 1. 
\resizebox{0.9\columnwidth}{!}{
$\begin{gathered}
    \\
    overlap = min(x_{end}, y_{end}) - max(x_{start}, y_{start})\\
    si = max\left(0, overlap\right) \\ 
    SCS = 1 - \left( \frac{si^2}{\text{length}(x) * \text{length}(y)}\right)
    \\
\end{gathered}$
}
\\

Thus, if the predicted role filler is an exact match for the gold role filler, the $SCS$ is 0. If there is some overlap between the spans, the $SCS$ is between 0 and 1 (not inclusive), and if there is no overlap between the spans, the $SCS$ is 1. The order of comparison of the spans doesn’t change the $SCS$ score for both modes.

As the absolute mode is less sensitive to changes in span indices as compared to the geometric mean, we chose geometric mean for our analysis, as tiny changes in index positions result in a bigger change in the $SCS$ score.

\item {\bf Alter Role} changes the role of a role filler to another role within the same template.

\item {\bf Remove Duplicate Role Filler} removes a role filler that is coreferent to an already matched role filler.

\item {\bf Remove Cross Template Spurious Role Filler} removes a role filler that would be correct if present in another template (in the same role).

\item {\bf Remove Spurious Role Filler} removes a role filler that has not been mentioned in any of the gold templates for a given document.

\item {\bf Introduce Role Filler} introduces a role filler that was not present in the predicted template but was required to be present in the matching gold template.

\item {\bf Remove Template} removes a predicted template that could not be matched to any gold template for a given document.

\item {\bf Introduce Template} introduces a template that can be matched to an unmatched gold template for a given document. \\
\end{itemize}

For a given document, all singleton {\bf Alter Span} and {\bf Alter Role} transformations, as well as sets of {\bf Alter Span + Alter Role} transformations, are applied first.

The other transformations are applied in the order in which they were detected, which is dependent on the order of predicted and gold template pairs in the optimized matching and the order of the slots/roles in the template.

\subsection{Error Type Mappings}
\label{subsec:error-type-mappings}
The transformations in Section~\ref{sec:transformation} are mapped onto a set of IE-specific error types as shown in Figure~\ref{fig:error_types}.  In some cases, a single transformation maps onto a single error, while in others a sequence of transformations is associated with a single error. Full details are in Appendix~\ref{appendix:mapping}.

\begin{table*}[t]
\small \centering
\begin{tabular}{l|c|c|c|c}
\toprule
      & \begin{tabular}[c]{@{}l@{}}\# docs \\ (train/val/test/unannot.)\end{tabular} & \begin{tabular}[c]{@{}l@{}}\# tokens per doc \\ (min/max/avg.)\end{tabular} & \begin{tabular}[c]{@{}l@{}}\# templates per \\ relevant doc (max/avg.)\end{tabular} & 
      \begin{tabular}[c]{@{}l@{}}\% docs with \\ 0 templates \end{tabular} \\ \midrule
\begin{tabular}[c]{@{}l@{}}- MUC-4 \\ \cite{muc-1992-message}\end{tabular}    &    1300 / 200 / 200 / 0                             &              31 / 1695 / 362                        &  14 / 1.61 & 44.59 \\
- ProMED\footnote{http://www.promedmail.org} &     125 / 12 / 108 / 4979                           &        57 / 4417 / 621                            &  9 / 1.55   &   19.83 \\
\begin{tabular}[c]{@{}l@{}}- SciREX \\ \cite{jain-etal-2020-scirex}\end{tabular}    &     304 / 66 / 66 / 0                            &    1153 / 13155 / 5401                                   &     16 / 2.28  & 0.00  \\ \bottomrule
\end{tabular}
\caption{Dataset Statistics. A relevant document has one or more templates.}
\label{tab:stats2}
\end{table*}

\section{Document-level IE Datasets}

Our experiments employ three document-level information extraction datasets. We briefly describe each below. Dataset statistics are summarized in Table \ref{tab:stats2}.\\

\noindent
\textbf{MUC-4}~\cite{muc-1992-message} consists of newswire describing terrorist incidents in Latin America provided by the FBIS (Federal Broadcast Information Services).  We converted the optional templates to required templates and removed the subtypes of the incidents as done in previous work \cite{chambers-2013-event, du-etal-2021-template} so that the dataset is transformed into standardized templates. The roles chosen from the MUC-4 dataset are {\sc PerpInd} (individual perpetrator), {\sc PerpOrg} (organization perpetrator), {\sc Target} (physical target), {\sc Victim} (human target), and {\sc Weapon} which are all string-fill roles, as well as {\sc incident type} which is a set-fill role with six possible role fillers: {\tt attack}, {\tt kidnapping}, {\tt bombing}, {\tt arson}, {\tt robbery}, and {\tt forced work stoppage}. As seen in Table \ref{tab:stats2},
44.59\% of the documents have no templates, which makes the identification of relevant vs.\ irrelevant documents critical to the success of any IE model for this dataset.\\

\noindent
\textbf{ProMED}\footnote{\url{http://www.promedmail.org}}~\cite{patwardhan-riloff-2009-unified} consists of just 125 annotated tuning examples and 120 annotated test examples, describing global disease outbreaks by subject matter experts from ProMED. We use the tuning data as training data and reserve 10\% of the test data, i.e.\ 12 examples, to create a development/validation set. 19.83\% of the documents in the dataset have no templates. The roles that we extract from the dataset are {\sc Status}, {\sc Country}, {\sc Disease}, and {\sc Victims}. {\sc Disease}, {\sc Victims}, and {\sc Country} are string-fill roles\footnote{In the ProMED dataset, {\sc Country} is a set-fill role, but since countries are explicitly mentioned in most of the documents, we can treat this role as a string-fill.}; {\sc Status} is a set-fill role with {\tt confirmed}, {\tt possible}, and {\tt suspected} as the possible role filler options.\\ 
\noindent
\textbf{SciREX}~\cite{jain-etal-2020-scirex} consists of annotated computer science articles from Papers with Code\footnote{\url{https://paperswithcode.com}}. We focus specifically on its 4-ary relation extraction subtask. The roles present in each relation are {\sc Material (Dataset)}, {\sc Metric}, {\sc Task}, and {\sc Method} which are all string-fills.  We convert the dataset from its original format to templates for our models, and remove individual role fillers (entities) that have no mentions in the text.\footnote{According to Jain et.\ al., around 50\% of relations in the SciREX dataset contain one or more role fillers that do not appear in the corresponding text. These relations are removed during evaluation for our end-to-end task. https://github.com/allenai/SciREX/blob/master/README.md} 
We also remove any duplicate templates.\footnote{Removing unmentioned entities sometimes eliminates differences between templates. This results in some templates becoming identical or making some templates contain information that is a subset of the information present in another template. Thus, we only keep one of these processed templates.} 
During preprocessing, we remove malformed words longer than 25 characters, as the majority of these consist of concatenated words that are not present in the corresponding text.

\section{IE Modeling Details}

In our experiments, we train and test two neural-based IE models, described briefly below, on 
the MUC-4, ProMED, and SciREX datasets. Note that to create the training data for both the DyGIE++ and GTT models, we use the first mention of each role filler in the document as the mention to be extracted.

\paragraph{DyGIE++ with Clustering} We use DyGIE++ \\ \phantom{ }--- a span-based, sentence-level extraction model\\ \phantom{ }--- to identify role fillers in the document and associate them with certain role types.
During training, the maximum span length enumerated by the model is set to 8 tokens as in
\newcite{wadden-etal-2019-entity} for the SciREX dataset and 11 tokens for the ProMED dataset. We use {\it bert-base-cased} and {\it allenai/scibert\_scivocab\_uncased} for the base BERT and SciBERT models respectively, which both have a maximum input sequence length of 512 tokens.

To aggregate entities detected by DyGIE++ into templates, we use a clustering algorithm. For the SciREX dataset, we adopt a heuristic approach that assumes there is only one template per document, and in that template, we assign the named entities predicted by DyGIE++ for a document to the predicted role types. For the ProMED dataset, we use a different clustering heuristic that ensures that each template has exactly one role filler for the {\sc Country} and {\sc Disease} roles, as detailed in the dataset annotation guidelines. Also, since {\sc Status} has the value {\tt confirmed} in the majority of the templates, every template predicted has its {\sc Status} assigned as {\tt confirmed}. 

\paragraph{GTT} is an end-to-end document-level template-generating model. For the MUC-4 and SciREX datasets, GTT is run for 20 epochs, while for ProMED it is run for 36 epochs, to adjust for the smaller size of the dataset. All other hyper-parameters are set as in \newcite{du-etal-2021-template}. We use the same BERT and SciBERT base models as described in the DyGIE++ architecture above, both with a maximum input sequence length of 512 tokens.

The computational budget and optimal hyperparameters for these models can be found in Appendix sections \ref{appendix:comp} and \ref{appendix:hyper}, respectively.

\section{Experimental Results and Analysis}

We first discuss the results of DyGIE++ and GTT on SciREX, ProMED, and MUC-4; and then examine the performance of these newer neural models on the 1992 MUC-4 dataset  vs.\ a few of the best-performing IE systems at the time.

\subsection{DyGIE++ vs.\ GTT}

Table~\ref{tab:f1} shows the results of evaluating DyGIE++ and GTT on the SciREX, ProMED, and MUC-4 datasets.  We can see that

{\bf all models perform substantially worse on scientific texts (ProMED, SciREX) as compared to news (MUC-4)}, likely because the model base is pretrained for general-purpose NLP applications (BERT) or there are not enough examples of scientific-style text in the pretraining corpus (SciBERT). 

In addition, models seem to perform better on the news-style ProMED dataset than the scientific-paper-based long-text SciREX dataset. This can be explained by the fact that all four models handle a maximum of 512 tokens as inputs, while the average length of a SciREX document is 5401 tokens. Thus, a majority of the text is truncated and, hence, unavailable to the models.

Nevertheless, we see an increase in F1 scores for all SciBERT-based models when compared to their BERT counterparts for the SciREX dataset. The same trend is seen for DyGIE++ for ProMED, but 
not for GTT.

This can be explained by the fact that GTT (SciBERT) has more Missing Template errors than
GTT (BERT). So even if GTT (SciBERT) performs better on the scientific slot {\sc Victims}, 
i.e.\ it extracts more scientific information, it does not identify relevant events as well as GTT (BERT), reducing the F1 score across the remaining slots.

\begin{figure*}[ht]
\centering
\resizebox{\textwidth}{!}{
\includegraphics[width=6.5in]{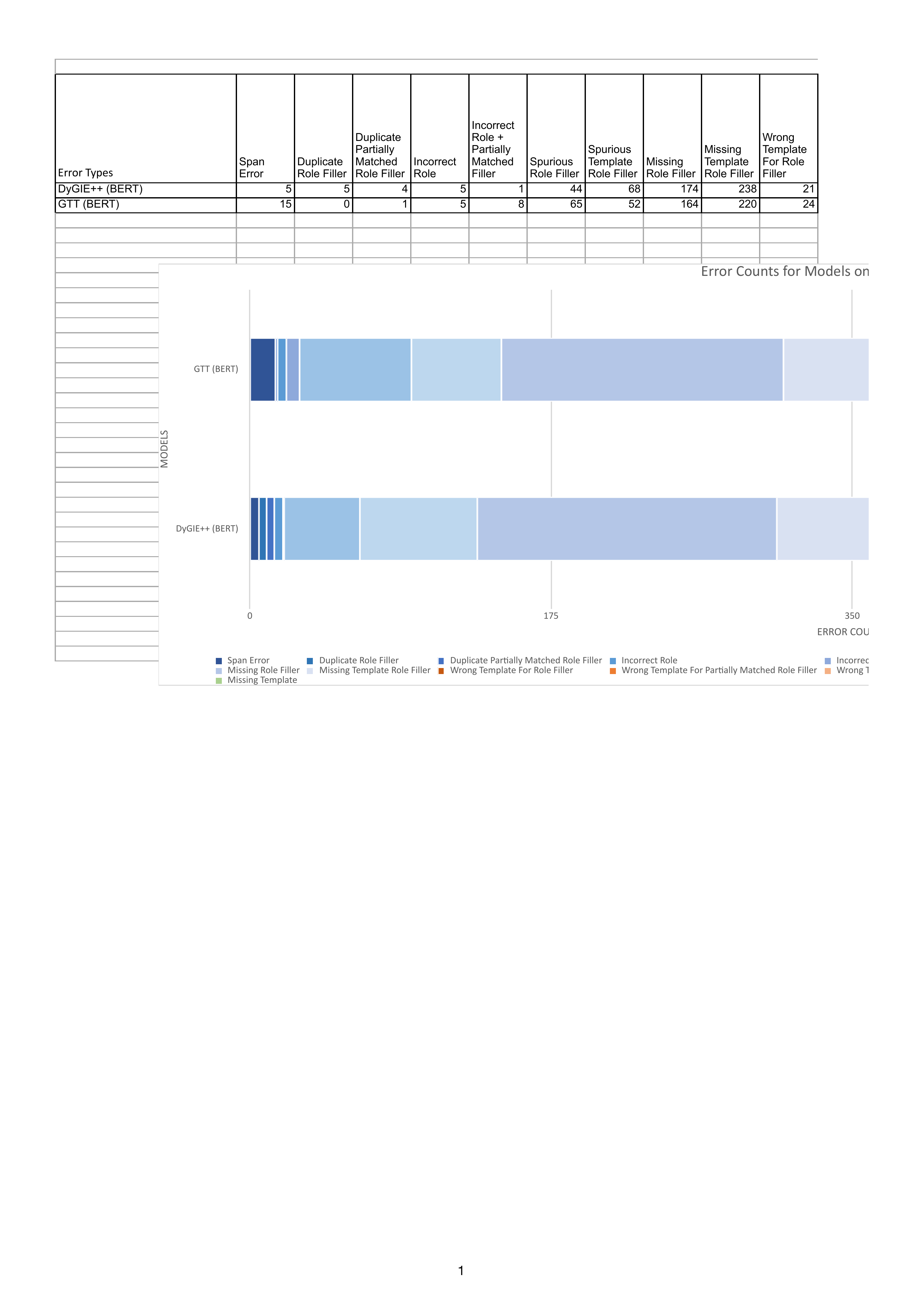}
}

\caption{Automated Error Analysis Results (Error Counts) for Models on the MUC-4 dataset.}

\label{fig:autoerrmuc}
\end{figure*}

From the error count results in Figure~\ref{fig:autoerrmuc}, we see that {\bf GTT makes fewer Missing Template errors than DyGIE++ on the MUC-4 dataset} (86 vs.\ 97). However, there is no significant difference in the number of missing templates between the two models on the ProMED and SciREX datasets. This could be because DyGIE++ is prone to overgeneration --- there are significantly more Spurious Role Filler and Spurious Template errors as compared to the results of GTT. Since we use heuristics that create templates based on the extracted role fillers, this increases the probability that there was a possible match to a gold template, reducing the number of Missing Template Errors.
 
We can also conclude that {\bf DyGIE++ is worse at coreference resolution when compared to GTT} as DyGIE++ makes more Duplicate Role Filler errors across all datasets.

Overall, we find that {\bf the major source of error for both GTT and DyGIE++ across all the datasets is missing recall} in the form of Missing Role Filler and Missing Template errors.

\begin{table}[!t]
\small \centering
\begin{tabular}{l|ccc}
\toprule
               & SciREX & ProMED & MUC-4 \\ \midrule
DyGIE++ (BERT) &  22.47\%     &  35.01\%      &    45.79\%   \\
DyGIE++ (SciBERT) &  25.39\%     &  38.15\%      &    -  \\
GTT (BERT)     &  21.54\%      &   {\bf 44.64}\%     &    {\bf 49.00\%}   \\
GTT (SciBERT)  &  {\bf 27.68}\%      &   42.96\%    &   -   \\ \bottomrule
\end{tabular}
\caption{F1 Scores for the Neural Models on SciREX, ProMED, and MUC-4}
\label{tab:f1}
\end{table}

\begin{table}[!t]
\small \centering
\begin{tabular}{l|ccc}
\toprule
               & Precision & Recall & F1 \\ \midrule
GE NLToolset &  56.69\%     &  {\bf 52.09\%}      &    {\bf54.29\%}   \\
NYU PROTEUS     &  34.23\%      &   31.28\%     &    32.69\%   \\
SRI FASTUS  &  48.47\%      &   38.42\%    &   42.86\%   \\ 
UMass CIRCUS  &  48.62\%      &   39.04\%    &   43.30\%  \\
\midrule
GTT (BERT) & {\bf 63.18}\% & 40.02\% & 49.00\% \\
DyGIE++ (BERT) & 61.90\% & 36.33\% & 45.79\%\\
\bottomrule
\end{tabular}
\caption{Precision, Recall, and F1 scores for models on the MUC-4 dataset. The first four models were developed in 1992, while the last two models are recent and use neural-based methods.}
\label{tab:progressmuc}
\end{table}

\begin{table}[!t]
\small \centering
\begin{tabular}{c|c}
\toprule
               Predicted & Gold Match \\ \midrule
\parbox[t]{0.2\textwidth}{power lines {\bf along the road}} & \parbox[t]{0.2\textwidth}{power lines}   \\ & \\
\parbox[t]{0.2\textwidth}{enrique ruiz{\bf, retired}} & \parbox[t]{0.2\textwidth}{enrique ruiz}  \\ & \\
\parbox[t]{0.2\textwidth}{{\bf maoist} shining path group} & \parbox[t]{0.2\textwidth}{shining path}  \\ & \\
 \parbox[t]{0.2\textwidth}{
group of unidentified individuals {\bf who hurled a bomb ... passing vehicle}} & \parbox[t]{0.2\textwidth}{group of unidentified individuals}  \\ & \\
\bottomrule
\end{tabular}
\caption{Span Errors in early models. The differences between the predicted mention and its best gold mention match according to our analysis tool are in bold.}
\label{tab:progressspan}
\end{table}

\subsection{Early IE Models vs.\ DyGIE++ and GTT}

Table~\ref{tab:progressmuc} presents the precision, recall, and F1 performance on the MUC-4 dataset 
for early models from 1992 alongside those of the more recent DyGIE++ and GTT models.  
We summarize key findings below.

\paragraph{The best of the early models (GE NLToolset) performs better than either of the modern models.} It does so by doing a better job balancing precision and recall, whereas GTT and DyGIE++ exhibit much higher precision and much lower recall.

\paragraph{The early models have more span errors than the modern DyGIE++ and GTT models.} The representative kinds of span errors from the 1992 model outputs are shown in Table~\ref{tab:progressspan}. One interesting difference between the span errors in the early models and the modern models is that the predicted mentions include longer spans with more information than is indicated in the best gold mention match. Some could be due to errors in dataset annotation; for example, {\it maoist shining path group} versus {\it shining path} but a significant number of the span errors occur as the early models seem to extract the entire sentence or clause which contains the desired role filler mention. The modern models tend to leave off parts of the desired spans, and if they do predict larger spans than required, are only off by a few words.
\paragraph{The early models have fewer Missing Template and Missing Role Filler errors as compared to the modern models.} However, the former also have more Spurious Template and Spurious Role Filler errors than the latter, indicating these models mitigate the issue of Missing Templates through over-generation. 
\paragraph{The early models have fewer Incorrect Role errors as compared to modern models.} However, since all the models make relatively few such errors, it suggests that role classification for predicted mentions is not a major problem for modern models. 
\paragraph{The main source of error for both early and modern models is missing recall due to missing templates and missing role fillers.} This strongly suggests future systems can maximize their performance by being less conservative in role filler detection and focusing on improvement of the recall, even at the expense of potentially decreasing some precision.

\vspace*{-3mm}
\section{Limitations and Future Work}

\vspace{-2mm}
This work explores subtypes of Spurious Role Filler errors extensively, however, we would like to further analyze Missing Role Filler and template-level errors for more fine-grained error subtypes and the linguistic reasons behind why they occur.

Due to the pairwise comparisons between all predicted and gold mentions in a role for all pairs of predicted and gold templates in an example, the error analysis tool is slow when the number of both the predicted and gold templates as well as the number of role fillers in the templates is high. Thus, we would also like to improve the time complexity of our template (and mention) matching algorithms using an approach like bipartite matching \cite{yang-etal-2021-document}. 

Currently, the error analysis tool reports exact match precision/recall/F1 which is more suitable for string-fill roles. We would like to extend the tool to further analyze set-fill roles by implementing metrics such as false-positive rate.

We used a limited number of models in this paper as we aimed to develop and test the usability of our error analysis tool. In the future, however, we would like to test our tool on a wider range of models, in addition to running more experiments in order to reach more generalizable conclusions about the behavior of IE models.

\section{Conclusion}
As new models for information extraction continue to be developed, we find that their
predicted error types contain  insights regarding their shortcomings. Analyzing error patterns within model  predictions in a more fine-grained manner beyond scores provided by commonly used metrics 

is important for the progress of the field.
We introduce a framework for the automatic categorization of model prediction errors for
document-level

IE tasks.   We used the tool to analyze the errors of two state-of-the-art models
on three datasets from varying domains and compared the error profiles of these models to four of the 
earliest systems in the field on a dataset from that era.
We find that state-of-the-art models, when compared to the earlier manual feature-based models, perform better at span extraction but worse at template detection and role assignment. With a better balance between precision and recall, the best early model outperforms the relatively high-precision, low-recall modern models. Missing role fillers remain the main source of errors, and scientific corpora are the most difficult for all systems, suggesting that improvements in these areas should be a priority for future system development. 

\section*{Acknowledgments}

We thank the anonymous reviewers and Ellen Riloff for their helpful comments(!) and Sienna Hu for converting the 1992 model outputs to a format compatible with our error analysis tool. Our research was supported, in part, by NSF CISE Grant 1815455 and the Cornell CS Department CSURP grants for undergraduate research.
 \newpage

\bibliographystyle{acl_natbib}

\bibliography{final_ref}

\clearpage

\appendix
\onecolumn

\section{Detailed Error Types Mappings}
\label{appendix:mapping}
The specific list of transformations applied in the error correction process. \\

(1) {\bf Span Error.} Each singleton Alter Span transformation is mapped to a Span Error. A Span Error occurs when a predicted role filler becomes an exact match to the a gold role filer only upon span alteration.

(2) {\bf Duplicate Role Filler.} Each singleton Remove Duplicate Role Filler transformation is mapped to a Duplicate Role Filler error. A Duplicate Role Filler error occurs when a spurious role filler is coreferent to an already matched role filler and is treated as a separate entity. This happens when the system fails at coreference resolution.

(3) {\bf Duplicate Partially Matched Role Filler (Spurious).} Same as (2) above, but with an added Alter Span transformation applied first to account for partial matching. This happens when the system fails at correct span extraction and coreference resolution.

(4) {\bf Spurious Role Filler.} Each singleton Remove Spurious Role Filler transformation is mapped to a Spurious Role Filler error. A Spurious Role Filler error occurs when a mention is extracted from the text with no connection to the gold templates.

(5) {\bf Missing Role Filler.} Each singleton Introduce Role Filler transformation is mapped to a Missing Role Filler error. A Missing Role Filler error occurs when a role filler is present in the gold template but not the predicted template for a given role.

(6) {\bf Incorrect Role.} Each singleton Alter Role transformation is mapped to an Incorrect Role. An Incorrect Role occurs when a spurious role filler is assigned to the incorrect role within the same template, i.e.\ the role filler would have been correct if present filled in another slot/role in the same template. This happens when the system fails at correct role assignment.

(7) {\bf  Incorrect Role + Partially Matched Filler.} Same as (4) above, but with an added Alter Span transformation applied first to account for partial matching. This happens when the system fails at correct span extraction and role assignment.

(8) {\bf Wrong Template for Role Filler.} Each singleton Remove Cross Template Spurious Role Filler transformation is mapped to a Wrong Template for Role Filler error. A Wrong Template for Role Filler occurs when a spurious role filler in one template can be assigned to the correct role in another template, i.e.\ it would be correct if it had been placed in another template. This happens when the system fails at correct event assignment.

(9) {\bf Wrong Template for Partially Matched Role Filler.} Same as (6) above, but with an added Alter Span transformation applied first to account for partial matching. This happens when the system fails at correct span extraction and event assignment.

(10) {\bf Wrong Template + Wrong Role.} An Alter Role and Remove Cross Template Spurious Role Filler transformation are applied to the same predicted role filler in that order to be mapped to a Wrong Template + Wrong Role error. A Wrong Template + Wrong Role error occurs when a spurious role filler can be assigned to another role in another template. This happens when the system fails at correct role assignment and event assignment.

(11) {\bf Wrong Template + Wrong Role + Partially Matched Filler.} Same as (8) above, but with an added Alter Span transformation applied first to account for partial matching. This happens when the system fails at correct span extraction, role assignment and event assignment.

(12) {\bf Spurious Template.}\footnote{The role fillers in the Spurious Templates are not added to the Spurious Role Filler error counts but are accounted for in the Spurious Template Role Filler counts.} Each singleton Remove Template is mapped to a Spurious Template error. A Spurious Template error occurs when an extra predicted template is present that cannot be matched to a gold template.

(13) {\bf Missing Template.}\footnote{The role fillers in the Missing Templates are not added to the Missing Role Filler error counts but are accounted for in the Missing Template Role Filler counts.} Each singleton Introduce  Template transformation is mapped to a Missing Template error. A Missing Template error occurs when there is a gold template remaining that has no matching predicted template.

\section{Example Error Types with ProMED}

We also provide example error types with the ProMED dataset.
\begin{table*}[h]
\centering
\small
\resizebox{\textwidth}{!}{
\begin{tabular}{|p{0.05\textwidth}|p{0.2\textwidth}|p{0.2\textwidth}|p{0.25\textwidth}|p{0.25\textwidth}|}
\hline
\multicolumn{1}{|c|}{} &
\multicolumn{1}{c|}{{\bf Error Types}}                         & \multicolumn{1}{c|}{{\bf Transformations(s)}}                              & \multicolumn{1}{c|}{{\bf Predicted}} & \multicolumn{1}{c|}{{\bf Gold}} \\ \hline
i) & Span Error & Alter Span & Victims: {\bf[\color{black}young fattening cattle]} & Victims: [young fattening cattle and sheep] \\ \hline
ii) & Duplicate Role Filler & Remove Duplicate Role Filler & Disease: [west nile fever], {\bf [\color{black}west nile virus]} & Disease: [west nile fever, west nile virus] \\ \hline
iii) & Within Template Incorrect Role & Alter Role & \parbox[t]{0.2\textwidth}{{\bf \color{torange} T1:} \\ Disease: {\bf [\color{black} 2 humans]} \\ Victims: ---} & \parbox[t]{0.2\textwidth}{{\bf \color{torange} T1:} \\ Disease: --- \\ Victims: [2 humans]} \\ \hline
iv) & Wrong Template For Role Filler & Remove Cross Template Spurious Role Filler & \parbox[t]{0.2\textwidth}{{ \bf \color{torange} T1:} \\ Country: [netherlands] \\ Victims: {\bf \color{black} [770 cases]}} &  \parbox[t]{0.2\textwidth}{{\bf \color{torange} T1:} \\ Country: [netherlands] \\ Victims: [its 11th case]} \\ 
& & & & \parbox[t]{0.2\textwidth}{{\bf \color{tpurple} T2:} \\ Country: [united kingdom] \\ Victims: [770 cases] } \\
\hline
v) & Spurious Template & Remove Spurious Template & \parbox[t]{0.2\textwidth}{{\bf \color{black}T1: Country: [china] }} & \multicolumn{1}{c|}{---} \\ \hline
vi) & Missing Template & Introduce Missing Template &  \multicolumn{1}{c|}{{\bf \color{black}---}} & \parbox[t]{0.2\textwidth}{{\bf \color{torange} T1}: \\ Country: [germany] \\Disease: [fmd] \\ Victims: [2 pigs]} \\ \hline
\end{tabular}}
\caption{Some examples of the Error Types taken from the ProMED dataset. For each template, in every role, the role fillers within brackets refer to the same entity, while role fillers in different brackets refer to different entities. The text in bold black indicates the error in the prediction.}
\label{tab:errexpromed}
\end{table*}

\section{Precision, Recall, and F1 Scores for All Models on all Three Datasets}
We also provide additional precision, recall scores along with the F1 scores.
\begin{table}[h]
\small \centering
\begin{tabular}{l|c|c|c}
\toprule
\multicolumn{1}{c|}{Models} & SciREX & ProMED & MUC-4 \\ \midrule
DyGIE++ (BERT) & 27.85 / 18.83 / 22.47
 & 51.13 / 26.62 / 35.01 & 61.90 / 36.33 / 45.79 \\
DyGIE++ (SciBERT) & 30.47 / 21.76 / 25.39 & 52.55 / 29.94 / 38.15 & - \\
GTT (BERT) & 52.86 / 13.53 / 21.54 & 68.58 / 33.09 / {\bf 44.64} & 63.18 / 40.02 / {\bf 49.00} \\
GTT (SciBERT) & 53.68 / 18.65 / {\bf 27.68} & 64.68 / 32.16 / 42.96 & - \\ 
\bottomrule
\end{tabular}
\caption{Precision, Recall and F1 Scores (\%).}
\label{tab:modelstats}
\end{table}

\section{Computational Budget}
\label{appendix:comp}
The GTT (BERT) model on the MUC-4 dataset took 1 hour and 21 minutes to train and around 11 minutes to test on Google Colab (GPU).

The GTT (BERT) model on the ProMED dataset took around 24 minutes to train and 4 minutes to test, while the GTT (SciBERT) model on the ProMED dataset took around 13 minutes to train and 4 minutes to test, both on Google Colab (GPU). The DyGIE++ (BERT) model on the ProMED dataset took around 50 minutes to train, while the DyGIE++ (SciBERT) model on the ProMED dataset took around 1 hour and 30 minutes to train, both on a NVIDIA V100 GPU.

For the SciREX dataset, it took around 10-20 minutes to run the GTT (BERT) and GTT (SciBERT) models on a NVIDIA V100 GPU. It is worth noting that since the GTT model embeds all inputs before training and SciREX documents are extremely long, more than 25 GB of memory needs to be allocated at the embedding phrase. The training process has normal memory usage. The DyGIE++ (BERT) model took around 2 hours to train, while the DyGIE++ (SciBERT) model took around 4 hours to train, both on a NVIDIA V100 GPU.

Our error analysis tool can be run completely on a CPU and takes a couple of minutes to run, depending on the size of the dataset and the predicted outputs.
\newpage
\section{Hyperparameters and Model Configurations}
\label{appendix:hyper}
We did not run the DyGIE++ model on the MUC-4 dataset as the model output was made available to us by Xinya Du. 

\begin{table}[!h]
\small \centering
\begin{tabular}{c|c}
     & GTT (BERT)  \\
    \toprule
    Hyperparameter Name & Value \\ 
    \midrule
    number of gpus & 	1 \\
    number of tpu cores & 	0\\
    max\_grad\_norm & 	1.0\\
    gradient\_accumulation\_steps & 	1\\
    seed&	1\\
    base\_model& bert\_base\_uncased\\
    learning\_rate &	5e-05\\
    weight\_decay &	0.0\\
    adam\_epsilon &	1e-08\\
    warmup\_steps &	0\\
    num\_train\_epochs &	20\\
    train\_batch\_size &	1\\
    eval\_batch\_size &	1\\
    max\_seq\_length\_src &	435\\
    max\_seq\_length\_tgt &	75\\
    threshold	& 80.0\\
    \bottomrule
\end{tabular}
\caption{GTT on the MUC-4 dataset}
\end{table}

\begin{table}[!h]
\small \centering
\begin{tabular}{c|c|c}
     & GTT (BERT) & GTT (SciBERT)\\
    \toprule
    Hyperparameter Name & Value & Value\\ 
    \midrule
    number of GPUs & 	1 & 	1\\
    number of TPU cores & 	0 & 	0\\
    max\_grad\_norm & 	1.0 & 	1.0\\
    gradient\_accumulation\_steps & 	1& 	1\\
    seed&	1&	1\\
    base\_model& bert\_base\_uncased & allenai\_ scibert\_ \\
    &  & scivocab\_uncased\\
    learning\_rate &	5e-05&	5e-05\\
    weight\_decay &	0.0 &	0.0\\
    adam\_epsilon &	1e-08 &	1e-08\\
    warmup\_steps &	0 &	0\\
    num\_train\_epochs &	36&	36\\
    train\_batch\_size &	1&	1\\
    eval\_batch\_size &	1&	1\\
    max\_seq\_length\_src &	435&	435\\
    max\_seq\_length\_tgt &	75 &	75\\
    threshold	& 80.0& 80.0 \\
    \bottomrule
\end{tabular}
\caption{GTT Models on the ProMED dataset}
\end{table}

\begin{table}[!h]
\small \centering
\begin{tabular}{c|c|c}
     & GTT (BERT) & GTT (SciBERT)\\
    \toprule
    Hyperparameter Name & Value & Value\\ 
    \midrule
    number of GPUs & 	1 & 	1\\
    number of TPU cores & 	0 & 	0\\
    max\_grad\_norm & 	1.0 & 	1.0\\
    gradient\_accumulation\_steps & 	1& 	1\\
    seed&	1&	1\\
    base\_model& bert\_base\_uncased & allenai\_ scibert\_ \\
    &  & scivocab\_uncased\\
    learning\_rate &	5e-05&	5e-05\\
    weight\_decay &	0.0 &	0.0\\
    adam\_epsilon &	1e-08 &	1e-08\\
    warmup\_steps &	0 &	0\\
    num\_train\_epochs &	20&	20\\
    train\_batch\_size &	1&	1\\
    eval\_batch\_size &	1&	1\\
    max\_seq\_length\_src &	435&	435\\
    max\_seq\_length\_tgt &	75 &	75\\
    threshold	& 80.0& 80.0 \\
    \bottomrule
\end{tabular}
\caption{GTT Models on the SciREX dataset}
\end{table}

\begin{table}[!h]
\small \centering
\begin{tabular}{c|c|c}
     & DyGIE++ (BERT) & DyGIE++ (SciBERT)\\
    \toprule
    Hyperparameter Name & Value & Value\\ 
    \midrule
    number of GPUs & 	1 & 	1\\
    max\_span\_width & 11 & 11\\
    base\_model& bert\_base\_cased & allenai\_ scibert\_ \\
    &  & scivocab\_cased\\
    learning\_rate &	5e-04&	5e-04\\
    patience &	5 &	5\\
    num\_train\_epochs &	20&	20\\
    train\_batch\_size &	32&	32\\
    num\_dataloader\_workers & 2 & 2\\
    max seq length &	512&	512\\
    ner loss weight& 1.0 & 1.0\\
    relation loss weight& 0.0 & 0.0\\
    coreference loss weight& 0.2 & 0.2\\
    events loss weight& 0.0 & 0.0\\
    target task & ner & ner\\
    \bottomrule
\end{tabular}
\caption{DyGIE++ Models on the ProMED dataset}
\end{table}

\begin{table}[!h]
\small \centering
\begin{tabular}{c|c|c}
     & DyGIE++ (BERT) & DyGIE++ (SciBERT)\\
    \toprule
    Hyperparameter Name & Value & Value\\ 
    \midrule
    number of GPUs & 	1 & 	1\\
    max\_span\_width & 8 & 8\\
    base\_model& bert\_base\_cased & allenai\_ scibert\_ \\
    &  & scivocab\_cased\\
    learning\_rate &	5e-04&	5e-04\\
    patience &	5 &	5\\
    num\_train\_epochs &	20&	20\\
    train\_batch\_size &	32&	32\\
    num\_dataloader\_workers & 2 & 2\\
    max seq length &	512&	512\\
    ner loss weight& 1.0 & 1.0\\
    relation loss weight& 0.0 & 0.0\\
    coreference loss weight& 0.2 & 0.2\\
    events loss weight& 0.0 & 0.0\\
    target task & ner & ner\\
    \bottomrule
\end{tabular}
\caption{DyGIE++ Models on the SciREX dataset}
\end{table}
\end{document}